\documentclass[a4paper,conference]{IEEEtran}

\usepackage{graphicx}

\ifCLASSINFOpdf
\else
\fi

\usepackage{amsmath}
\usepackage{bm}
\usepackage{textcomp}
\usepackage{multirow}
\usepackage{float}
\usepackage{stfloats}
\usepackage{hyperref}

\usepackage[ruled,vlined]{algorithm2e}

\begin{document}
%

\title{Smart Inference for Multidigit Convolutional Neural Network based Barcode Decoding}

\author{\IEEEauthorblockN{Thao Do}
\IEEEauthorblockA{KAIST\\Daejeon, South Korea\\
thaodo@kaist.ac.kr}
\and
\IEEEauthorblockN{Yalew Tolcha}
\IEEEauthorblockA{KAIST\\Daejeon, South Korea\\
yalewkidane@kaist.ac.kr}
\and
\IEEEauthorblockN{Tae Joon Jun}
\IEEEauthorblockA{Asan Medical Center \\Seoul, South Korea\\
taejoon@amc.seoul.kr}
\and
\IEEEauthorblockN{Daeyoung Kim}
\IEEEauthorblockA{KAIST\\Daejeon, South Korea \\
kimd@kaist.ac.kr}}

\maketitle

\begin{abstract}
Barcodes are ubiquitous and have been used in most critical daily activities for decades. However, most traditional decoders require well-founded barcode under a relatively standard condition. While wilder conditioned barcodes such as underexposed, occluded, blurry, wrinkled and rotated are commonly captured in reality, those traditional decoders show weakness of recognizing. Several works attempted to solve those challenging barcodes, but many limitations still exist. This work aims to solve the decoding problem using deep convolutional neural network with the possibility of running on portable devices. Firstly, we proposed a special modification of inference based on the feature of having checksum and test-time augmentation, named Smart Inference (SI), in the prediction phase of a trained model. SI considerably boosts accuracy and reduces the false prediction for trained models. Secondly, we have created a large practical evaluation dataset of real captured 1D barcode under various challenging conditions to test our methods vigorously, publicly available for other researchers. The experiments' results demonstrated the SI effectiveness with the highest accuracy of 95.85\% which outperformed many existing decoders on the evaluation set. Finally, we successfully minimized the best model by knowledge distillation to a shallow model which is shown to have high accuracy (90.85\%) with a good inference speed of 34.2 ms per image on a real edge device.
\end{abstract}


\IEEEpeerreviewmaketitle

\section{Introduction}
Linear 1D barcodes appeared in the 1970s and are now become ubiquitous on almost all consumer products and for logistics due to its ease of identification. Some newer tagging technologies emerged over the last decades allowing more information (e.g. RFID, NFC) stored. However, none of them has fully replaced its role in the industry because of its legacy and its economy. The low cost of printing barcode and the durability of the tag under minor damages make it remain an industry standard (standardized by GS1) for the coming decades.

One essential property of the tagging technology is that it must be read quickly, robustly and accurately using readers. For the barcode case, the readers (or the scanners) are categorized into 3 types: laser-based, LED-based, and camera-based. In the first 2 types, the laser/LED ray needs to be close to the barcode, requires no stripe obscured on the ray line, and suffers the problem of emitter overheating. Camera-based readers have some advantages over laser/LED-based solutions. The first advantage is built on the fact that numerous smartphones with high-quality cameras integrated are already in use. With Internet connection, useful mobile applications were born by online retrieval of product information and giving out ingredients information, alerting allergies, calorie intake, comparing prices between sellers; or for retailers, they learn eye-catching products, have consumer feedback and so on (e.g. in \cite{fernandcz2017image}). Another advantage of the camera-based solution is the possibility of multiple and long-range recognition by the support of computer vision algorithms.

\begin{figure*}[!ht]
\centering
  \includegraphics[width=\textwidth,height=8cm,keepaspectratio]{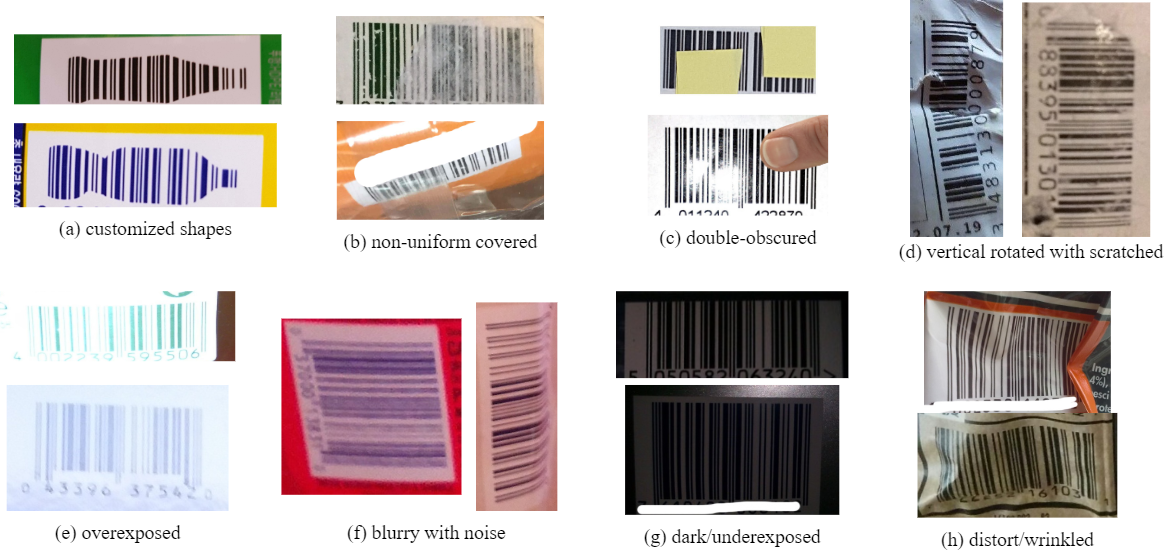}
  \caption{Challenging conditions}
  \label{fig:fig1}
\end{figure*}

However, most current techniques (static image processing and pattern matching) being used in camera-based readers have flaws that limit their usability. The main problem with them is the need for well-framed flatbed-scanned-style input than normal captured. Wilder but common-captured conditions such as underexposed, occluded, blurry or curved, non-horizontal position (as in Figure \ref{fig:fig1}) become unrecognizable. This requires the user correction which is unhandy and slows down the scanning process. There are 2 separate tasks in scanning barcode: detecting (i.e. locating) where the barcode region in the image and decoding detected region to barcode sequence. Recent works showed that the first task is nearly solved even in challenging conditions.

On the other hand, decoding those challenging barcodes still needs to improve since existing works still have many limitations. Traditional methods presented in \cite{joseph1994bar, muniz1999robust, wachenfeld2008robust} apply traditional techniques like Hough transformations, scanline-based approach with thresholds for binarization based on certain assumptions of barcode characteristics while they are not always true. Many evaluated their tools on unpublished sets, some published their sets but small and not enough challenging conditions. With those limitations and the successes of convolutional neural network (CNN) in many applications, \cite{fridborn2017reading} was the first proposed work using CNN to decode these difficult codes. However, their work has some weak points making the performance much lower than CNN potential. Not only their CNN feature extractors are simple but also their input assumption is oversimplified. They only assumed the horizontal barcode as input; their test set is made from printed rectangle-shaped generated barcodes on plain papers while real-life barcode is printed in customized shapes (e.g. coca icon shape) on various materials with many kind of distortion and sometimes covered by film. They did not also consider the possibility of running the task on edge devices because of unoptimized models.

Therefore, in this study, we proposed a CNN-based method to solve decoding task with the following contributive points: (i) we proposed Smart Inference - 3 algorithms leveraging the feature of having checksum and test-time augmentation built on top of trained deep CNN models which considerably boost model accuracies and reduce false prediction; (ii) we made a challenging 2500-sample-cropped EAN13(UPC-A, ISBN13 are its subsets) barcode dataset from real captured images on various (included harsh) conditions on numerous products - this dataset is published for other researchers to evaluate their models and encourage more contribution on this task; (iii) lastly, we applied knowledge distillation technique with a target to have a lightweight model from the best model which is suitable on handheld devices, the experimental result consequently confirmed the possibility by a good inference speed on a real edge board.

\section{Related work}
Regarding the barcode locating task, there are some methods presented with improving performance over the past decade. In 2011, Lin et al \cite{lin2011real} presented the first multiple and rotation tolerated barcode recognition methods. This work focused more on detecting problem by using several image processing schemes such as Gaussian smoothing filtering to segment out barcode regions, enhanced the stripes, rotated the regions to horizontal angle and put into a decoder with voting. Although the method did well on lottery barcodes (printed on plain papers), it was still slow and didn't get high accuracy on a dataset of merchandise products which was unclear about the challenging level. Katona et al \cite{katona2012novel} in 2012 proposed a method using morphological operations to also segment out 1D and 2D barcode under blurry, noise, shear and various rotated conditions with good performance. Soros et al \cite{soros2013blur} continued dealing with blur using structure matrix and saturation from HSV color system to detect blurry barcodes better but with the expense of lowing speed in 2013. Recently, Creusot et al. \cite{creusot2016low} proposed a faster method for blurry barcodes based on Line Segment Detector after their previous work \cite{creusot2015real} using Maximal Stable Extremal Region shown sensitive to blur. In another way, Hansen \cite{hansen2017real} first tried to apply an object detection deep learning model (YOLO) on both 1D and 2D codes with the best bounding box detection rate.

\begin{figure*}[!b]
\centering
  \includegraphics[width=\textwidth,height=10cm,keepaspectratio]{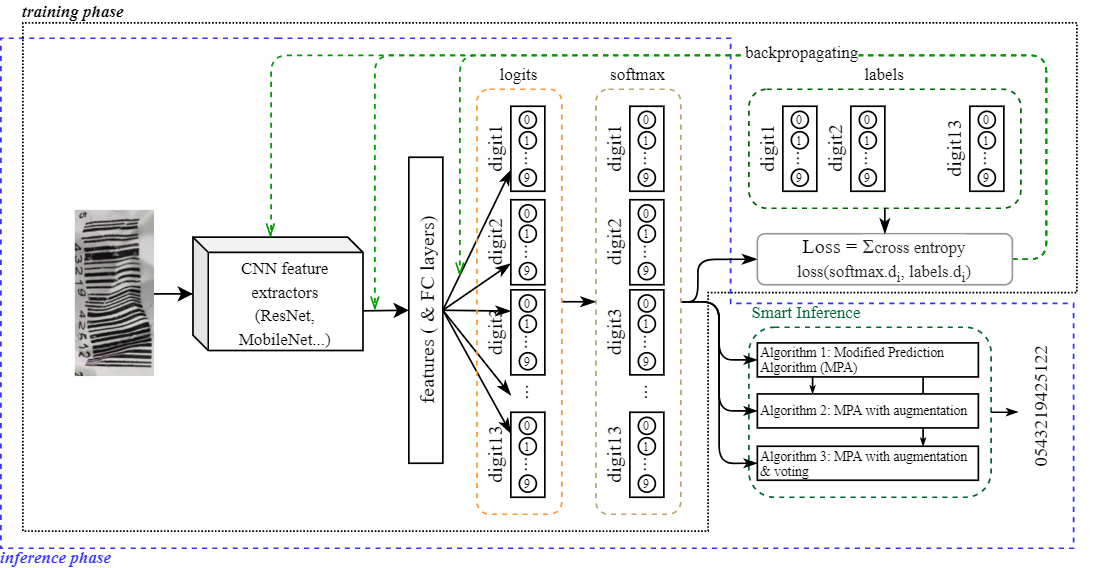}
  \caption{Overall Architecture}
  \label{fig:overall}
\end{figure*}

While the task of barcode detection nearly reached its saturation, several works on decoding had been proposed sparsely since 1990s. Early works \cite{joseph1994bar,muniz1999robust} achieved their goal by techniques as Hough transformation, wavelet-based peak location on their simple (scanned-style) inputs. Wachenfeld et al \cite{wachenfeld2008robust} proposed a scanline-based approach accompanied by an EAN13 dataset (so-called \emph{MuensterDB}). However, since their method was based on scanline approach at that time, it just worked well on slightly rotated (\textpm 15 degrees) barcode, stronger rotated or distorted would be problematic. Similar to \cite{katona2012novel}, Zamberletti et al \cite{zamberletti2010decoding} also tackled the problem of out-of-focus (blurry) barcode by using multilayer perceptron model to find parameters of adaptive thresholding (instead of standard binarization) to restore blurry image to clearer image and put it into Zxing \cite{owen2013zxing} to decode. Nonetheless, this approach is simple with low recall and time-consuming (2 steps). Recently, Yang et al \cite{yang2016automatic} tried to address 2 tasks on 5 rigorous datasets. The work outperformed all other methods very well on EAN13 barcodes, but since the method was heavily based on scanline-base and hand-crafty featuring and analysis for each challenging condition, it's a less scalable solution to extend the all other 1D barcode types (even though all 1D codes use stripes, they might differ in guard bar layouts, in how black-and-white stripe mixing) and the case of double-obscured condition (Figure \ref{fig:fig1}) would be inapplicable. Lastly, Fridborn \cite{fridborn2017reading} in 2017 first leveraged the power of CNN to directly extract features and predict to 13 outputs (correspond to 13 digits) simultaneously (similar to \cite{goodfellow2013multi} in Street View House Numbers problem). Compared with traditional and hand-crafty featuring methods, CNN-based approach is relatively more straightforward and data-driven rather than case-by-case analysis. One obvious example is the double-obscured condition which is problematic for scanline-based approaches but could be easily learnt and overcome by CNN classifier. Thus, our work is also CNN-based, however, differs from \cite{fridborn2017reading} by following points: (i) we use more advanced CNN feature extracting models; (ii) our input assumption is more practical as well as our training set and evaluation set covered more cases; (iii) we proposed Smart Inference exploiting the checksum attribute of barcode sequences to enhance model accuracies; (iv) we considered minimizing and verify the possibility of CNN-based approach on a real edge device.

Referring to test-time augmentation we used in this work for enhancing the inference accuracy a model, the technique is commonly used in deep learning as this survey \cite{shorten2019survey} and can be found in \emph{AlexNet} paper \cite{krizhevsky2012imagenet}, \emph{ResNet} paper \cite{he2016deep}. While train-time data augmentation gives more variants of the dataset to let the model also learn all possible variants; test-time augmentation also applies some proper modifications to original samples, let the model give multiple predictions on those modified versions and pick the most suitable one among these predictions by voting mechanism. How to augment data for better performance is also one trendy topic in deep learning now with such papers like \emph{AutoAugment} \cite{cubuk2019autoaugment}, \emph{Smart Augmentation} \cite{lemley2017smart}. In our work, we integrated test-time augmentation into Smart Inference quite effectively.

On the topic of model compression and deep learning applicability on mobile, Cheng et al \cite{cheng2017survey} categorized methods into 4 types: Parameter pruning and sharing, low-rank factorization, compact convolutional filters and knowledge distillation. The first one reduces redundant parameters which are not sensitive to the performance while the second one uses matrix decomposition to estimate the informative parameters. The third one builds special filters to save parameters for only convolutional layers whereas the last technique trains a compact neural network with knowledge distilled from a large model which is so-called teacher model. For simplicity, in this paper, we used the original knowledge distillation (KD) method proposed by Hinton et al \cite{hinton2015distilling}.

\section{Methodology}
The base approach we use in this work is to train a probabilistic model of decoding barcode sequences given barcode images as \cite{fridborn2017reading}. Let $\mathbf{D}$ represent the barcode sequence and $X$ represent the input barcode image. The goal is to learn a model of $P(\mathbf{D}|X)$ by maximizing $\log P(\mathbf{D}|X)$ on the training set. $\mathbf{D}$ is modelled as a collection of $n$ random variables $D_1,...,D_{n}$ representing $n$ digits of the decoded sequence. To simplify, assume that the value of the each digits are independent from each other, so that the probability of a sequence $d=d_1,...,d_n$ is given by $P(\mathbf{D}=\mathbf{d}|X)=\prod^{n}_{i=1}P(D_i=d_i|X)$. Each of the digits is discrete and has 10 possible values (0 to 9). This means each digit could be represented with a softmax classifier that receives as input features extracted from $X$ by a CNN. This type of model is originally proposed by \cite{goodfellow2013multi}, so we call it Multidigit CNN. During the training phase, the loss is calculated by the sum of all cross-entropy losses of digits as usual. However, in the inference phase, instead of normal inference, we propose a modification named Smart Inference (SI), which is one of the main contributions in this work. The detail of SI is described next paragraphs. The overall model is shown in Figure \ref{fig:overall}.

\begin{algorithm}
\SetAlgoLined
\SetKwInOut{Input}{Input}
\SetKwInOut{Output}{Output}
\Input{Trained Multidigit CNN (MDCNN), Barcode Image (BI), Maximum Iteration (Max), Voting status (Voting) }
\Output{Barcode Digit Combination/s (BDC)}
    Compute logit[ ][ ] using MDCNN given BI\;
 \For{$K\gets1$ \KwTo $N$}{
    $prob^K[\ ] \leftarrow softmax(logit^K[ \ ])$\;
	Descending  sorting  $Prob^K[\ ]$\;
    $diff^K \leftarrow Prob^K[0].val - Prob^K[1].val$\;
    $digit^K \leftarrow  Prob^K[0].index$\;
    append $gap\{=diff^k$\} and $position\{=K\}$ to $gap\_list$\;
 }
 Ascending sorting $gap\_list$  with element $gap$ \;
 $initial\_combination \leftarrow$  new combination with $digit[ ]$\;
 append $initial\_combination$ to $combination\_list$\;
 initialize $iter$ to zero
 \For{each $gap$ $\in$ gap\_list}{
    increment $iter$ by one\;
    \If{$iter$ is greater than $Max$ }{
            \uIf{voting }{
               \textbf{return} $voting\_combinations$\;
             }
             \Else{
             \textbf{return} null\;}
         }
    $K \leftarrow$ gap.position\;
    $new\_digit \leftarrow Prob^K[1].index$\;

    \For{each $combination \in combination\_list$}{
       $new\_combination \leftarrow$ modify $combination$ at position $K$ with $new\_digit$\; 
       append $new\_combination$ to $new\_combination\_list$\;
    }
    \For{each $combination \in new\_combination\_list$}{
       $status  \leftarrow$ Compute checksum test for $combination$\;
       \If{status }{
           \uIf{voting }{
                append $combination$ to $voting\_combinations$\;
             }
             \Else{
             \textbf{return} $combination$\;}
            
         }
    }
 }
 \textbf{return} $null$
\caption{Modified Prediction Algorithm (MPA)}
\label{MPA}
\end{algorithm}

\begin{algorithm}
\SetAlgoLined
\SetKwInOut{Input}{Input}
\SetKwInOut{Output}{Output}
\Input{Trained Multidigit CNN (MDCNN), Barcode Image (BI), Maximum Iteration (Max) }
\Output{Barcode Digit Combination (BDC)}
$degree\_list \leftarrow$ append degrees [90,180,270]\;
$image\_list \leftarrow$ rotate image BI with $degree\_list$\;
\For{each $image \in image\_list$}{
       $combination \leftarrow$ Prediction using \textbf{Algorithm \ref{MPA}} given MDCNN, $image$, and voting = False \;  
       \If{$combination$ is not null }{
            \textbf{return} $combination$\;
         }
    }
    \textbf{return} null 
\caption{MPA with Augmentation}
\label{MPA_Augmentation}
\end{algorithm}

\begin{algorithm}
\SetAlgoLined
\SetKwInOut{Input}{Input}
\SetKwInOut{Output}{Output}
\Input{Trained Multidigit CNN (MDCNN), Barcode Image (BI), Maximum Iteration (Max) }
\Output{Barcode Digit Combination (BDC)}
$degree\_list \leftarrow$ append degrees [90,180,270]\;
$image\_list \leftarrow$ rotate image BI with $degree\_list$\;
\For{each $Image \in Image\_list$}{
       $combinations \leftarrow$ Prediction with \textbf{Algorithm \ref{MPA}} given MDCNN, $image$, and voting = True \;
       \If{$combination$ is not null }{
         append  $combinations$ to  $combination\_list$\;
         }
}
\uIf{$combination\_list$ is not empty }{
        group combinations which are similar\;
         $combination \leftarrow$ select combination with highest count\;
         \textbf{return} $combination$\;
         }
         \Else{
   \textbf{return} null\;
  }
\caption{MPA with Augmentation and Voting}
\label{MPA_Augmentation_Voting}
\end{algorithm}

\subsection{Smart Inference}
Normally after getting logits from the model given barcode images, we apply softmax function to get the probabilities of each value (0 to 9) for all $n$ digits; then, we pick the value with the highest probability. By this way, we finally have $n$-digit sequences from values having the highest probabilities. However, the value with the highest probability is not always the correct value. Instead, the correct value maybe the value with the second-highest or third-highest probability. Besides, since most 1D barcodes have a characteristic of checksum satisfaction as \cite{checkdigit}. Let $D$ is the barcode sequence, $D[i]$ is the digit $i^{th}$ of the sequence from left to right, $L$ is the length of the barcode sequence (e.g. length of EAN13 is 13) (so first digit is $D[1]$), the checksum attribute could be summarized as this equation:

\begin{equation} \label{checksum_equation}
\begin{split}
(D[L-0]*\mathbf{1} + D[L-1]*\mathbf{3} + \\
D[L-2]*\mathbf{1} + D[L-3]*\mathbf{3} +...+ \\
D[L-2i]*\mathbf{1} + D[L-2i-1]*\mathbf{3} +...+ \\ 
D[1]*\mathbf{1} \: \text{if} \: L \bmod 2 == 0 \: \text{else} \: \mathbf{3}) \bmod 10 = 0
\end{split}
\end{equation}

Leverage this characteristic, our initial idea was to make more than one predicted sequence from not only the value with the highest probability but also from value having $2^{nd}$ probability (or $3^{rd}$) highest for each digit of $n$ digits; then, we verify those combinations by equation (\ref{checksum_equation}). Intuitively, the bigger gap between the value having the highest probability and the value having $2^{nd}$ (or $3^{rd}$) highest probability is , the more confident the model predicts the value having the highest probability and vice versa. Therefore, it is a priority to consider those digits having the smallest gaps where the model is more confused and less certain in only value having the highest probability. Let $V$ is the number of values having the highest probabilities and \emph{Maxim Iteration} is the number of digits considered more than one value (as in Algorithm \ref{MPA}). In this work, because the bigger $V$ or \emph{Maxim Iteration} is, the larger number of combination created causes inference downtime, we only picked $V$ = 2 (i.e. we only consider 2 values having 2 highest probabilities), and conduct experiments with \emph{Maxim Iteration} from $1$ to $4$ (i.e. the value of each of ($n-\emph{Maxim Iteration}$) other digits is the value having highest probability, we have $2^\emph{Maxim Iteration}$ combinations). Lastly, we sort candidate combinations by order of larger to smaller probability and test the equation satisfaction (\ref{checksum_equation}) one by one, stop at the first satisfying combination for fast inference. This process is clearly described in Algorithm \ref{MPA}.

The Algorithm \ref{MPA} is enhanced by applying test-time augmentation in 2 ways: fast-track as in Algorithm \ref{MPA_Augmentation} and voting as Algorithm \ref{MPA_Augmentation_Voting}. For simplicity and fast inference which is important in this application, we only used 3 rotation operations to augment each input image. Algorithm \ref{MPA_Augmentation} iterates through original input and 3 its variants step-by-step calling Algorithm \ref{MPA} and stops as soon as Algorithm \ref{MPA} gets the first equation satisfying combination, otherwise, no decoded sequence is returned. On the other hand, Algorithm \ref{MPA_Augmentation_Voting} collects satisfying combinations from all iterations (original input \& variants) and picks the most frequent combinations.

One thing we need to emphasize this idea compared to \cite{yang2016automatic} is our proposed techniques are more scalable for other types of 1D barcodes (EAN, UPC, ITF barcode family) with a few changes. This technique could be applicable for multiple barcode types in one model, we just need to add a few more nodes, some to categorize barcode types, some to fill up the length of the longest barcode types (each digit now having 11 values, 0-9 and NA), the equation \ref{checksum_equation} still applicable to all other EAN, UPC codes.

\subsection{Minimize Deep Model}
To minimize deep models to have a more suitable model for edge devices, we use original knowledge distillation technique in \cite{hinton2015distilling} to distill knowledge from the best (deep) model to small shallow models by replacing original loss function by combined loss:
\[ L = (1 - \alpha)*L_H + \alpha*L_{KL} \]

Where $L_H$ is the cross-entropy loss from the hard labels, $L_{KL}$ is the Kullback–Leibler divergence loss from the teacher labels (soft label) and $\alpha$ is hyperparameter.

\section{Experiments}
\subsection{Datasets}

Our real collected set is comprised of 1055 samples from extended \emph{MuensterDB}, 408 samples from \cite{zamberletti2010decoding} and 1037 our self-collected. Totally we have 2500 samples after we drew bounding box, labeled the decoded sequences (\cite{wachenfeld2008robust} and \cite{zamberletti2010decoding} had not finished both tasks in their datasets). Our self-collected samples are captured from 5 supermarkets (1 in France, 2 in South Korea, 2 in Vietnam) both indoors and outdoors for 2 weeks. A wide range of products from food and edible product packages, books, kitchenware, office stationery, clothes tags on various material such as metal cans, wine bottles, food plastic bags, cardboard box under various light sources (florescent light, incandescent bulb, morning and afternoon sunlight) and conditions (auto-focus off, handshaking, long-distance, obscured by fingers, wrinkled, distort, cornered); also 195 printed barcodes on plain papers with occluded and wrinkled conditions. This set is available at \emph{\href{https://www.resl.kaist.ac.kr/doc/datasets}{resl.kaist.ac.kr/doc/datasets}}

\begin{table}[H]
\caption{Synthesized conditions}
\label{tab:synthesize-conditions}
\begin{tabular}{ll}
\emph{Condition(s)}                      &\emph{Number of samples}\\
\hline
norm                                     & 30000             \\
dark                                     & 30000             \\
occluded                                 & 20000             \\
occluded+dark                            & 20000             \\
rotated \& perspective transformed (RPT) & 20000             \\
RPT + dark                               & 20000             \\
cylindered \& curvy warped (CCW)         & 20000             \\
CCW + dark                               & 20000             \\
occluded + RPT                           & 5000              \\
blur                                     & 5000              \\
RPT + blur                               & 5000              \\
CCW + blur                               & 5000              \\
upside down                              & 6000              \\
upside down + dark                       & 6000              \\
upside down + blur                       & 6000              \\
upside down + CCW                        & 6000              \\
upside down + occluded                   & 6000              \\
heavy noise + rotated                    & 2000              \\
overexposed + occluded + RPT + CCW       & 6000              \\
dark + occluded + RPT + CCW              & 6000              \\
occluded + RPT + CCW                     & 6000             
\end{tabular}
\end{table}

\begin{figure}[H]
    \includegraphics[width=\linewidth,height=7cm,keepaspectratio]{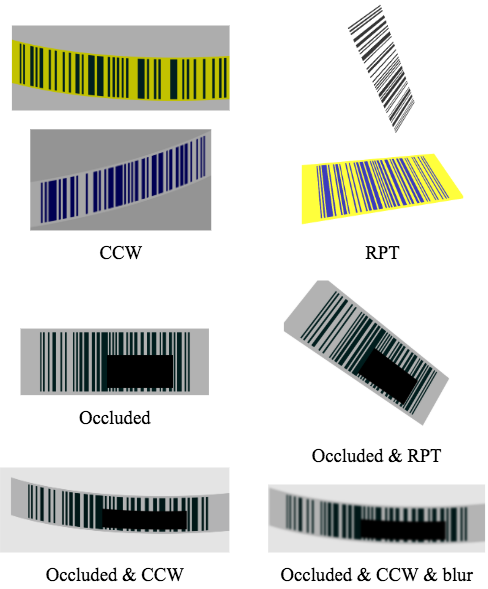}
    \centering
    \caption{Some Synthesized Samples}
    \label{fig:syn}
\end{figure}

Our training set consists of 250000 synthesized samples (without decoded text under stripes) with conditioned described in Table \ref{tab:synthesize-conditions} (note that 40000 samples are randomly added noise) and 20000 samples augmented from 500 real samples chosen randomly from the real collected set. Some of the synthesized samples are shown in Figure \ref{fig:syn}. The rest 2000 samples of the real collected set are used as test set.

\subsection{Experimental Setups}
To demonstrate our proposed model's performance improvement, we have done experiments on enterprise solutions and base models with deep learning. Zxing \cite{owen2013zxing} is an open-source tool used by most developers, while Google Barcode API is a commercialized version of Zxing. On the other hand, Cognex and Dynamsoft are two large corporations with a long history of developing products using machine vision for industrial uses. For these 4 tools, to make fair evaluations, we all applied test-time augmentation just the same way with our deep learning models in Algorithm \ref{MPA_Augmentation} and set them to work specifically for EAN13. Note that we used Zxing source and Google API latest versions, Dynamsoft and Cognex demo web-based API were used to evaluated so we also deduced round-trip message duration in measuring inference time. We already considered comparing our results with \cite{wachenfeld2008robust, lin2011real, zamberletti2010decoding,yang2016automatic} methods, but since we could not get any source or binary, we just stopped at using the listed public tools.

Regarding deep neural network models, Fridborn-similar model is similar to what described in their work \cite{fridborn2017reading}, (since their input was 196x100x1 while ours is 285x285x3, so that 4 convolutional blocks results in exceeding GPU resource) what we had to change are 32 kernels instead of 256 kernels for last convolutional layer and 2048 nodes instead of 4096 nodes for each of 2 top FC layers). Next, we modeled non-residual model with 8 convolutional blocks and 2 FC layers having many fewer parameters compared to Fridborn-similar one. Other models using SOTA CNN feature extractors such as ResNet50, ResNet34 \cite{he2016deep}, MobileNetV2 \cite{sandler2018mobilenetv2}, DenseNet169 \cite{huang2017densely} just have original feature extractor parts come directly before 13 output nodes as in Figure \ref{fig:overall}. Various batch sizes were tried but in our empirical observation, 32 might be the best number. All models were trained from scratch without pretrained knowledge. Note that we had to train models with only synthesized set first until the loss reduce to around 1 (i.e. models converged to a certain level) before training on full training set (with 20000 real-collected augmented samples) because directly training on full training set results in a very high loss (even NaN).
The training processes were made using NVIDIA Titan RTX with 24 GB VRAM. Our CPU evaluation experiments were conducted on desktop using Intel Core i9 9900KF processor, 32GB RAM while low-computational experiments were run on a CUDA-enabled NVIDIA Jetson Nano board using NVIDIA TensorRT models (converted from PyTorch).

\subsection{Evaluation}
We have 2 metrics to clarify here: accuracy and errors. Basically, a tool would have 3 outcome states given a barcode image: correct (i.e. match the ground truth) decoded sequence, incorrect decoded sequence and no barcode existed (or no checksum-satisfied sequence for our proposed models). Accuracy metric in this work is calculated by 
\[ Accuracy = \frac{\# \: of \: correct \: decoded \:sequences}{\# \: of \: total \: barcode \: images} \]
while the number of error = $\# \: of \: the \: incorrects$. This means a good model is the one that achieves higher accuracy and fewer errors. Another figure that needs to be mentioned in this section is the inference time which is average inference time per one image since each image takes a different amount of time for decoders.

\begin{table}[ht]
\caption{Tool \& models without MPA performances}
\label{tab:result-basic}
\begin{tabular}{llll}
\emph{Model}& \emph{Accuracy} & \emph{CPU (ms)} & \emph{\# of params (M)} \\
\hline
Zxing            & 58.25\%  & \textbf{7.65}     & NA               \\
Dynamsoft        & 93.10\%  & 978.8    & NA               \\
Google API       & 82.45\%  & 211.9    & NA               \\
Cognex           & 84.60\%  & 111.9    & NA               \\
\hline
ResNet50         & \textbf{93.35\%}  & 66.5     & 99.5             \\
MobiletNetV2     & 72.25\%  & 32.4     & \textbf{15.7}             \\
MobiletNetV2\_kd & 83.45\%  & 32.4     & \textbf{15.7}             \\
DenseNet169      & 84.90\%  & 75.65    & 30               \\
ResNet34         & 88.70\%  & 38.3     & 40.7             \\
ResNet34\_kd     & 89.20\%  & 38.3     & 40.7             \\
Fridborn-similar & 31.85\%  & 104.9    & 403.3            \\
Non-residual     & 80.80\%  & 103.2    & 78.5            
\end{tabular}
\end{table}

\begin{table}[ht]
\caption{Using MPA performances}
\label{tab:mpa}
\begin{tabular}{llllllll}
    & \emph{Model} & \emph{nonMPA} & \emph{max=1} & \emph{max=2} & \emph{max=3} & \emph{max=4}\\
\hline
\parbox[t]{1mm}{\multirow{5}{*}{\rotatebox[origin=c]{90}{Accuracy}}}& ResNet50& 0.9335& 0.942& 0.9435& \textbf{0.9445}& 0.9445\\
                      & MobiletNetV2 & 0.8345 & 0.8595 & \textbf{0.869}  & 0.868  & 0.8645\\
                      & DenseNet169      & 0.849  & 0.8645 & 0.876  & \textbf{0.8775} & 0.874 \\
                      & ResNet34     & 0.892  & 0.9075 & 0.911  &\textbf{0.912}  & 0.911 \\
                      & Non-residual     & 0.808  & 0.8295 & 0.844  & \textbf{0.8455} & 0.841 \\
\hline
\parbox[t]{1mm}{\multirow{5}{*}{\rotatebox[origin=c]{90}{\# of errors}}}& ResNet50& 133 & \textbf{31} & 48 & 77 & 106\\
                      & MobiletNetV2 & 331    & \textbf{59}     & 106    & 164    & 241\\
                      & DenseNet169      & 302    & \textbf{54}     & 105    & 176    & 230\\
                      & ResNet34     & 216    & \textbf{41}     & 73     & 116    & 157\\
                      & Non-residual     & 384    & \textbf{55}     & 104    & 197    & 285
\end{tabular}
\end{table}

\begin{table}[ht]
\caption{Using MPA \& Augmentation performances}
\label{tab:aug}
\begin{tabular}{lllllll}
    & \emph{Model} & \emph{nonMPA} & \emph{max=1} & \emph{max=2} & \emph{max=3} & \emph{max=4}\\
\hline
\parbox[t]{1mm}{\multirow{5}{*}{\rotatebox[origin=c]{90}{Accuracy}}}& ResNet50& 0.9335& \textbf{0.958}& 0.956  & 0.951  & 0.946  \\
                        & MobiletNetV2 & 0.8345 & \textbf{0.906}  & 0.8975 & 0.8855 & 0.8705 \\
                        & DenseNet169  & 0.849  & \textbf{0.9155} & 0.9075 & 0.8915 & 0.877  \\
                        & ResNet34     & 0.892  & \textbf{0.9375} & 0.9315 & 0.9235 & 0.916  \\
                        & Non-residual & 0.808  & \textbf{0.89}   & 0.879  & 0.861  & 0.8445 \\
\hline
\parbox[t]{1mm}{\multirow{5}{*}{\rotatebox[origin=c]{90}{\# of errors}}}& ResNet50& 133 & \textbf{56} & 73 & 95 & 108    \\
                        & MobiletNetV2 & 331    & \textbf{121}    & 182    & 225    & 259    \\
                        & DenseNet169  & 302    & \textbf{113}    & 172    & 216    & 246    \\
                        & ResNet34     & 216    & \textbf{95}     & 129    & 152    & 168    \\
                        & Non-residual & 384    & \textbf{145}    & 212    & 274    & 311   
\end{tabular}
\end{table}

\begin{table}[!ht]
\caption{Using MPA \& Augmentation with Voting performances}
\label{tab:aug-vote}
\begin{tabular}{lllllll}
    & \emph{Model} & \emph{nonMPA} & \emph{max=1} & \emph{max=2} & \emph{max=3} & \emph{max=4}\\
\hline
\parbox[t]{1mm}{\multirow{5}{*}{\rotatebox[origin=c]{90}{Accuracy}}}& ResNet50& 0.9335& \textbf{0.9585}& 0.9585 & 0.9525 & 0.95\\
                        & MobiletNetV2 & 0.8345 & \textbf{0.9085} & 0.906  & 0.899  & 0.8945 \\
                        & DenseNet169  & 0.849  & \textbf{0.9125} & 0.8985 & 0.8785 & 0.8545 \\
                        & ResNet34     & 0.892  & \textbf{0.933}  & 0.93   & 0.9215 & 0.911  \\
                        & Non-residual & 0.808  & \textbf{0.8855} & 0.8735 & 0.853  & 0.8355 \\
\hline
\parbox[t]{1mm}{\multirow{5}{*}{\rotatebox[origin=c]{90}{\# of errors}}}& ResNet50 & 133 & \textbf{55} & 68 & 92 & 100\\
                        & MobiletNetV2 & 331    & \textbf{116}    & 165    & 198    & 211    \\
                        & DenseNet169  & 302    & \textbf{119}    & 190    & 242    & 291    \\
                        & ResNet34     & 216    & \textbf{104}    & 132    & 156    & 178    \\
                        & Non-residual & 384    & \textbf{154}    & 223    & 290    & 329   
\end{tabular}
\end{table}

Regarding inference time, we should note that it is hard to have good perfect evaluation since Dynamsoft, Google API and Cognex were tested via APIs which are run on their own servers that are not matched our configured desktop. Since some of the tools are not using machine learning (except Dynamsoft, Cognex use DNN on their many other products so theirs might be DNN model also), their processing time is relatively smaller than deep learning-based techniques but with low accuracy of prediction. The basic evaluation is presented in Table \ref{tab:result-basic}. Our base models outperform other models with reasonable computation time for prediction with accuracy more than 0.93.

To show the performance gained by applying Smart Inference during testing time, we have performed three different experiments. The first experiment is done with Algorithm \ref{MPA}. As depicted in Table \ref{tab:mpa}, the result shows that the performance is improved compared to models without it. The result also shows that the performance improves when the number of gaps considered is increased up to some level, it then shows degradation. The second experiment is done to demonstrate how (Algorithm \ref{MPA_Augmentation}) fast-track augmentation can improve MPA performance. It clearly shows a performance improvement over Algorithm \ref{MPA} as depicted in Table \ref{tab:aug} and significant improvements compared to basic approach. However, this time, just after considering one pair having the smallest gap, we already reached the best results. Like MPA without augmentation (Algorithm \ref{MPA}), the number of errors in Algorithm \ref{MPA_Augmentation} increases as the number of considering gap is increased. Nevertheless, Algorithm \ref{MPA} still has shown a smaller number of errors when compared to predictions with Algorithm \ref{MPA_Augmentation}. The third experiment corresponded to Algorithm \ref{MPA_Augmentation_Voting} which is done with more cost. As shown in table \ref{tab:aug-vote}. It sometimes slightly outperforms the experiments based on Algorithm \ref{MPA_Augmentation} with similar behavior when we change parameter $Max$. This suggests that the voting scenario is not always a good choice for our models and the models might already be relatively robust to the original input image and only need help after they failed in the first place.

As we mentioned in the last section, to demonstrate the possibility of the solution on portable devices, we distilled knowledge from the best model (ResNet50) to 2 considered small models: ResNet34 and MobileNetV2. The result in Table \ref{tab:result-basic} clearly shows that knowledge distillation does help in gaining higher performance compared with training by the normal loss function. MobileNetV2 jumps considerable from \emph{72.25\%} to \emph{83.45\%} while the improvement in ResNet34 model is not much. This could be because ResNet34 is still deep (40.7 million parameters compared to 15.7 million parameters of MobileNetV2) and so their learning ability from itself is still robust enough to reach high performance without the guidance from the teacher model. Finally, our experimental tests on the NVIDIA Jetson Nano board show that MobileNetV2, ResNet34 achieved average speeds of \emph{34.2}, \emph{45.6} milliseconds per images respectively. This speed is equivalent to a smooth frame-per-second experience with the robustness of the model, we expect it is comfortable for users.

\section{Conclusion}

In this work, we have proposed Smart Inference for Multidigit CNN based models to improve the performance of 1D barcode decoding. We have collected multiple real barcodes with label data to train and test the proposed model. We have also added better synthesized data to strengthen the training and testing process. The algorithms proposed during testing time boosted the performance over the base models. It not only outperforms the base model in terms of accuracy but also has small inference time which makes it efficient. The Multidigit CNN based approach with Smart Inference is also a scalable solution as it could extend to decode more than one barcode type. We have also shown that distillation technique transfers effectively knowledge from the best model to the shallower model to run on low computational edge devices and performs clearly better than training with normal loss function.

Even though the performance is better in terms of accuracy, the proposed model has a limitation in predicting false records (Dynamsoft also predicts 3 errors). Another limitation of this approach is that it is not applicable for non-fixed length barcode types such as Code39. In the future, the problem of false predicting can be mitigated by applying product recognition techniques.

\section{Acknowledgement}

This work was supported and funded by the Ministry of Science and ICT (MSIT) under the Korea-EU Joint Research Support Project of National Research Foundation of Korea (NRF-2016K1A3A7A0395205414), Main Research Program (E0162502) of the Korea Food Research Institute(KFRI), and Grand Information Technology Research Center support program (IITP-2020-0-01489) supervised by the IITP (Institute for Information \& communications Technology Planning \& Evaluation).

\bibliographystyle{IEEEtran}
\bibliography{IEEEabrv,IEEEexample}

\end{document}